\documentclass[runningheads]{llncs}
\usepackage[T1]{fontenc}
\usepackage{graphicx}
\usepackage{etoolbox}
\AtBeginDocument{%
  \setlength{\abovedisplayskip}{6pt plus 2pt minus 2pt}
  \setlength{\belowdisplayskip}{6pt plus 2pt minus 2pt}
  \setlength{\abovedisplayshortskip}{3pt plus 1pt minus 1pt}
  \setlength{\belowdisplayshortskip}{3pt plus 1pt minus 1pt}
}
\usepackage{dsfont}
\usepackage{amsmath}
\usepackage{caption}
\usepackage{tikz}
\usetikzlibrary{shapes.geometric, arrows.meta, positioning, fit, calc, decorations.pathreplacing}
\usepackage{booktabs}
\usepackage{xcolor}
\usepackage{pgfplots}
\pgfplotsset{compat=1.17}

\begin{document}

\title{Scaling State-Space Models from Lines to Paragraphs: An Ablation of Mamba-based OCR}

\author{Merveilles AGBETI-MESSAN\inst{1}\orcidID{0009-0007-5024-2138} \and
Thierry PAQUET\inst{1}\orcidID{0000-0002-2044-7542} \and
Clément CHATELAIN\inst{2}\orcidID{0000-0001-8377-0630}  \and
Pierrick TRANOUEZ\inst{1}\orcidID{0000-0002-1962-0782}  \and
Stéphane NICOLAS\inst{1}\orcidID{0000-0003-0575-6731}}

 \authorrunning{M. AGBETI-MESSAN et al.}
%
\institute{
LITIS EA4108, University of Rouen Normandy, France \\
\email{\{komlan-epe-nsin.agbeti-messan, thierry.paquet, pierrick.tranouez, stephane.nicolas\}@univ-rouen.fr}
\and
LITIS EA4108, INSA of Rouen Normandy, France \\
\email{\{clement.chatelain\}@insa-rouen.fr}
}

\titlerunning{Scaling Mamba from Lines to Paragraphs}
\maketitle

\begin{abstract}
End-to-end OCR increasingly relies on autoregressive sequence models, where the quadratic cost of Transformer attention limits efficient transcription of long, paragraph-level text. State-Space Models (SSMs) such as Mamba offer linear-time decoding and have recently been shown to match Transformer accuracy on printed historical lines, but their behavior as sequences grow from short lines to full paragraphs, and their generalization to handwriting, remain poorly understood. 

We study how a Mamba-based OCR recognizer scales from lines to paragraphs. We first conduct a systematic exploration of its four core hyperparameters (decoder depth, state dimension, expansion factor, and connector depth) on synthetic paragraphs from 100 to 1{,}000 characters, identifying the recurrent state dimension and the expansion factor as the dominant levers for long-sequence accuracy. We then compare the recognizer against a Transformer baseline trained under an identical protocol. On clean synthetic paragraphs, both models stay below 1\% CER at every length while the SSM runs 1.4 to 4.5 times faster, the speedup growing with sequence length. On real handwriting, however, the SSM lags clearly behind: it reaches 8.2\% CER on IAM lines and 10.0\% on IAM paragraphs, against 4.2\% and 3.5\% for the Transformer baseline. Through controlled experiments we show that a substantial part of this gap stems from data scarcity rather than from an intrinsic architectural limit: the autoregressive SSM decoder is markedly data-hungry on long sequences. 

Our study clarifies when SSMs are a practical choice for large-scale document transcription and when they are not.
\end{abstract}

\section{Introduction}
\label{sec:introduction}

Cultural heritage institutions worldwide face the task of digitizing millions of historical documents \cite{muehlberger2019transkribus,impact2011dataset}. Transformer-based architectures \cite{coquenet2023dan,li2021trocr} dominate current OCR research but incur $O(n^2)$ attention complexity, which limits practical deployment for long paragraph-level sequences. A growing body of work has explored linear-time alternatives: retention-based recurrent backbones such as RetNet \cite{sun2023retnet}, gated linear-recurrent variants such as Gated DeltaNet \cite{yang2024gateddeltanet}, and selective State-Space Models (SSMs) such as Mamba \cite{gu2023mamba}. Mamba combines content-adaptive state propagation with hardware-aware parallel scanning, while RetNet has very recently been shown to support strong handwritten text recognition through decoder-only designs such as DRetHTR \cite{kim2026drethtr}.

Recent work by Agbeti-Messan et al.~\cite{agbeti2026benchmark} introduced the first application of SSMs to historical document OCR, showing that Mamba-based recognizers can match Transformer accuracy on line-level recognition while providing substantial efficiency gains: 2.9$\times$ faster inference on Antiqua lines (53.3 vs 156.5 ms) and 2.05$\times$ on paragraphs (195.6 vs 401.2 ms), with empirically linear memory scaling versus Transformers' quadratic scaling.

Three questions remain open. First, the optimal architectural configuration for paragraph-level Mamba-OCR has not been justified: \cite{agbeti2026benchmark} used reasonable defaults (4 decoder layers, expansion factor 6, single bidirectional connector) without exploring the hyperparameter space. Second, the scaling behavior of Mamba-OCR has not been characterized in controlled conditions, where real document data confounds length effects with degradation, layout, and typographic variation. Third, the generalization of SSM-based OCR from printed text to handwriting remains unexplored.

This paper addresses these gaps. Building on \cite{agbeti2026benchmark}, we focus on the autoregressive variant (Mamba-AR), the only Mamba configuration capable of handling paragraph-level recognition in \cite{agbeti2026benchmark}. Our contributions are:

\begin{itemize}
    \item \textbf{Hyperparameter exploration:} We explore four key hyperparameters of Mamba-AR, namely decoder depth ($L \in \{2,4,8,16\}$), state dimension ($N \in \{16,32,64,128,256\}$), expansion factor ($E \in \{2,6,10,12,20\}$), and multimodal connector depth, at multiple sequence lengths (1, 3, 5, 7, 10 synthetic lines). Our results support the configuration of \cite{agbeti2026benchmark} and show that the state dimension is the only hyperparameter whose chosen value (256) is bounded by implementation constraints rather than by an accuracy plateau.
    \item \textbf{Controlled scaling study:} Using synthetic Wikipedia paragraphs (100 to 1000 characters), we characterize Mamba-OCR's behaviour with sequence length. We compare against DAN \cite{coquenet2023dan}, trained on the same data with the same protocol. Both models stay below 1\% CER at every length and neither dominates in accuracy, but Mamba-OCR is 1.4--4.5$\times$ faster, with the speedup growing with length.
    \item \textbf{Real-world validation and limitations:} We extend \cite{agbeti2026benchmark} (BnL lines and paragraphs) to handwritten data (IAM \cite{marti2002iam}). Mamba-OCR underperforms both Transformer baselines and DRetHTR \cite{kim2026drethtr} on IAM lines (8.21\% vs 2.26--4.23\% CER), and exhibits a large gap to DAN on IAM paragraphs (10.02\% vs 3.51\%). We show that the autoregressive Mamba decoder is data-hungry: 8K fixed paragraphs are insufficient for convergence even on short sequences, whereas 1M dynamically-sampled paragraphs enable stable training. This suggests that a substantial part of the IAM gap is attributable to data scarcity rather than to a purely architectural limitation.
\end{itemize}

The remainder of this paper is organized as follows. Section \ref{sec:background} recalls the Mamba-OCR architecture. Section \ref{sec:methodology} presents our synthetic data generation and training protocol. Section \ref{sec:ablation} reports the hyperparameter exploration. Section \ref{sec:scaling} presents the scaling curves. Section \ref{sec:realworld} validates findings on real data and analyzes limitations. Section \ref{sec:conclusion} concludes.

\section{Background}
\label{sec:background}

\subsection{Mamba-OCR Architecture}
\label{subsec:mamba-ocr-arch}

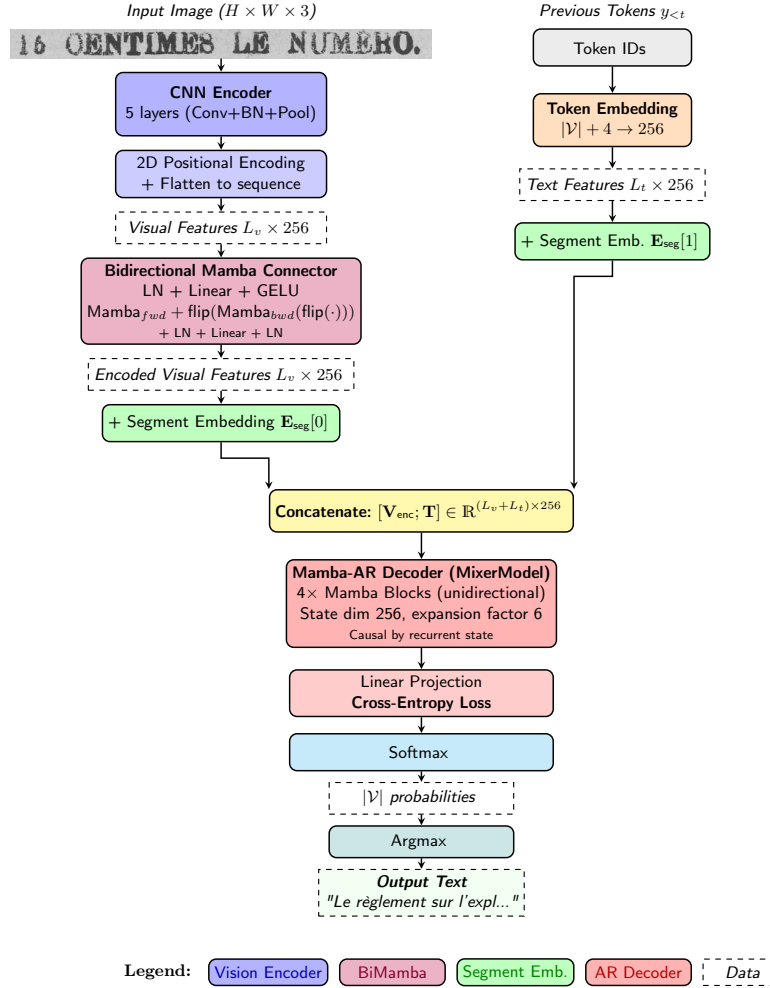
\begin{figure*}[!t]
\centering
\resizebox{0.85\textwidth}{!}{
\begin{tikzpicture}[
    node distance=0.7cm,
    procbox/.style={rectangle, draw, rounded corners, minimum width=3cm, minimum height=0.7cm, align=center, fill=#1, font=\sffamily\small, line width=0.8pt},
    procbox/.default={blue!20},
    bigprocbox/.style={rectangle, draw, rounded corners, minimum width=3cm, minimum height=1.2cm, align=center, fill=#1, font=\sffamily\small, line width=0.8pt},
    databox/.style={rectangle, draw, dashed, minimum width=3cm, minimum height=0.6cm, align=center, fill=white, font=\sffamily\small\itshape, line width=0.6pt},
    arrow/.style={->, >=stealth, thick},
    dashedarrow/.style={->, >=stealth, thick, dashed},
    label/.style={font=\sffamily\bfseries\footnotesize}
]

\node[font=\sffamily\small\itshape] (input_label) {Input Image ($H \times W \times 3$)};
\node[inner sep=0pt, below=0.05cm of input_label, minimum width=4cm] (input) {
    \includegraphics[width=8cm]{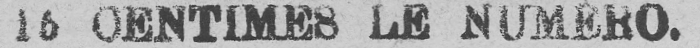}
};

\node[bigprocbox={blue!30}, below=0.25cm of input, minimum width=4cm] (encoder) {
    \textbf{CNN Encoder}\\
    5 layers (Conv+BN+Pool)
};

\node[procbox={blue!20}, below=0.25cm of encoder, minimum width=4cm] (pos2d) {
    2D Positional Encoding\\
    + Flatten to sequence
};

\node[databox, below=0.25cm of pos2d, minimum width=4cm] (features_flat) {
    Visual Features $L_v \times 256$
};

\node[bigprocbox={purple!30}, below=0.25cm of features_flat, minimum width=4cm, minimum height=1.4cm] (bimamba) {
    \textbf{Bidirectional Mamba Connector}\\
    LN + Linear + GELU\\
    $\text{Mamba}_{fwd} + \text{flip}(\text{Mamba}_{bwd}(\text{flip}(\cdot)))$\\
    \scriptsize{+ LN + Linear + LN}
};

\node[databox, below=0.25cm of bimamba, minimum width=4cm] (encoded) {
    Encoded Visual Features $L_v \times 256$
};

\node[procbox={green!25}, below=0.25cm of encoded, minimum width=4cm] (vis_seg) {
    + Segment Embedding $\mathbf{E}_{\text{seg}}[0]$
};

\node[font=\sffamily\small\itshape, right=4cm of input_label] (tok_label) {Previous Tokens $y_{<t}$};
\node[procbox={gray!15}, below=0.05cm of tok_label, minimum width=3cm] (tokens_in) {
    Token IDs
};

\node[procbox={orange!25}, below=0.5cm of tokens_in, minimum width=3cm, minimum height=1cm] (tok_emb) {
    \textbf{Token Embedding}\\
    $|\mathcal{V}|+4 \rightarrow 256$
};

\node[databox, below=0.4cm of tok_emb, minimum width=3cm] (text_feat) {
    Text Features $L_t \times 256$
};

\node[procbox={green!25}, below=0.4cm of text_feat, minimum width=3cm] (text_seg) {
    + Segment Emb. $\mathbf{E}_{\text{seg}}[1]$
};

\coordinate (merge_point) at ($(vis_seg.south) + (3.8cm, -1.3cm)$);
\node[procbox={yellow!40}, minimum width=5cm, minimum height=0.8cm] at (merge_point) (concat) {
    \textbf{Concatenate:} $[\mathbf{V}_{\text{enc}}; \mathbf{T}] \in \mathds{R}^{(L_v + L_t) \times 256}$
};

\node[bigprocbox={red!30}, below=0.5cm of concat, minimum width=5cm, minimum height=1.5cm] (decoder) {
    \textbf{Mamba-AR Decoder (MixerModel)}\\
    4$\times$ Mamba Blocks (unidirectional)\\
    State dim 256, expansion factor 6\\
    \scriptsize{Causal by recurrent state}
};

\node[procbox={red!20}, below=0.4cm of decoder, minimum width=5cm, minimum height=0.8cm] (proj) {
    Linear Projection\\
    \textbf{Cross-Entropy Loss}
};

\node[procbox={cyan!20}, below=0.3cm of proj, minimum width=5cm] (softmax) {
    Softmax
};

\node[databox, below=0.2cm of softmax, minimum width=3.5cm] (probs) {
    $|\mathcal{V}|$ probabilities
};
\node[procbox={teal!20}, below=0.2cm of probs, minimum width=3.5cm, minimum height=0.6cm] (argmax) {
    Argmax
};
\node[databox, below=0.2cm of argmax, minimum width=3.5cm, fill=green!5] (text) {
    \textbf{Output Text}\\
    "Le règlement sur l'expl..."
};

\draw[arrow] (input) -- (encoder);
\draw[arrow] (encoder) -- (pos2d);
\draw[arrow] (pos2d) -- (features_flat);
\draw[arrow] (features_flat) -- (bimamba);
\draw[arrow] (bimamba) -- (encoded);
\draw[arrow] (encoded) -- (vis_seg);

\draw[arrow] (tokens_in) -- (tok_emb);
\draw[arrow] (tok_emb) -- (text_feat);
\draw[arrow] (text_feat) -- (text_seg);

\draw[arrow] (vis_seg.south) -- ++(0, -0.3cm) -| (concat.north west);
\draw[arrow] (text_seg.south) -- ++(0, -0.3cm) -| (concat.north east);

\draw[arrow] (concat) -- (decoder);
\draw[arrow] (decoder) -- (proj);
\draw[arrow] (proj) -- (softmax);
\draw[arrow] (softmax) -- (probs);
\draw[arrow] (probs) -- (argmax);
\draw[arrow] (argmax) -- (text);

\node[below=0.8cm of text, xshift=-5cm] (legend_title) {\textbf{Legend:}};
\node[procbox={blue!30}, right=0.2cm of legend_title, minimum width=2cm, minimum height=0.5cm] (leg1) {\textcolor{blue!70!black}{Vision Encoder}};
\node[procbox={purple!30}, right=0.2cm of leg1, minimum width=2cm, minimum height=0.5cm] (leg2) {\textcolor{purple!70!black}{BiMamba}};
\node[procbox={green!25}, right=0.2cm of leg2, minimum width=2cm, minimum height=0.5cm] (leg3) {\textcolor{green!50!black}{Segment Emb.}};
\node[procbox={red!30}, right=0.2cm of leg3, minimum width=2cm, minimum height=0.5cm] (leg4) {\textcolor{red!70!black}{AR Decoder}};
\node[databox, right=0.2cm of leg4, minimum width=1.5cm, minimum height=0.5cm] (leg5) {Data};

\end{tikzpicture}
}
\caption{\textbf{Mamba-OCR (AR variant) architecture.} \textcolor{blue!70!black}{The shared CNN visual encoder} extracts features enriched with 2D sinusoidal positional encoding and flattened into a sequence. \textcolor{purple!70!black}{The bidirectional Mamba connector} models global visual context using a shared Mamba block processed forward and backward. \textcolor{green!50!black}{Segment embeddings} distinguish visual from text tokens before concatenation. \textcolor{red!70!black}{The autoregressive Mamba decoder} (4 unidirectional Mamba layers with state dimension 256 and expansion factor 6) processes the concatenated sequence and predicts characters through a linear projection. Causality is naturally enforced by Mamba's recurrent state, avoiding the need for attention masks.}
\label{fig:mamba-ar-architecture}
\end{figure*}

We build upon the Mamba-OCR architecture introduced in \cite{agbeti2026benchmark}, focusing exclusively on the autoregressive variant (Mamba-AR) which we simply denote \emph{Mamba-OCR} in this paper. Figure \ref{fig:mamba-ar-architecture} provides an architectural overview. The model comprises three main components: a shared CNN visual encoder, a bidirectional Mamba multimodal connector, and an autoregressive Mamba decoder.

\subsubsection{\textcolor{blue!70!black}{Shared CNN Encoder.}}
\label{subsubsec:cnn-encoder}
We adopt the CNN encoder from DAN \cite{coquenet2023dan}: five convolutional layers with batch normalization and max-pooling produce a 2D feature map of dimension $H' \times W' \times 256$, which is flattened to a sequence $\mathbf{V} \in \mathds{R}^{L_v \times 256}$ where $L_v = H' \times W'$. Before flattening, 2D sinusoidal positional encodings are added to preserve spatial information.

\subsubsection{\textcolor{purple!70!black}{Bidirectional Multimodal Connector.}}
\label{subsubsec:connector}
Since selective SSMs process sequences unidirectionally, a dedicated module is needed to provide bidirectional context. Our connector follows a residual architecture with normalization, projection, bidirectional Mamba processing, and a feedforward component. Formally, given the visual sequence $\mathbf{V} \in \mathds{R}^{L_v \times 256}$:
\begin{equation}
\tilde{\mathbf{V}} = \text{GELU}(\text{Linear}_1(\text{LayerNorm}_1(\mathbf{V})))
\end{equation}
\begin{equation}
\mathbf{h}^{\text{fwd}} = \text{Mamba}(\tilde{\mathbf{V}}), \quad \mathbf{h}^{\text{bwd}} = \text{flip}(\text{Mamba}(\text{flip}(\tilde{\mathbf{V}})))
\end{equation}
\begin{equation}
\mathbf{V}_{\text{enc}} = \text{LayerNorm}_3\big(\text{LayerNorm}_2(\mathbf{h}^{\text{fwd}} + \mathbf{h}^{\text{bwd}}) + \text{Linear}_2(\cdot)\big)
\end{equation}
where $\text{flip}(\cdot)$ reverses the temporal dimension. Crucially, the \emph{same} Mamba block is used for both directions (\emph{weight sharing}), reducing parameters while providing bidirectional context. The connector can optionally be stacked with multiple layers (briefly discussed in Section \ref{subsec:ablation-connector}); by default, a single layer is used.

\subsubsection{\textcolor{green!50!black}{Segment Embeddings.}}
\label{subsubsec:segment}
Since visual and text tokens are processed jointly by the decoder through concatenation, we use learned segment embeddings $\mathbf{E}_{\text{seg}} \in \mathds{R}^{2 \times 256}$ to mark each modality. The visual segment embedding $\mathbf{E}_{\text{seg}}[0]$ is added to all encoded visual positions, and the text segment embedding $\mathbf{E}_{\text{seg}}[1]$ is added to all token embeddings before concatenation.

\subsubsection{\textcolor{red!70!black}{Autoregressive Mamba Decoder.}}
\label{subsubsec:decoder}
The decoder follows the standard MixerModel architecture \cite{gu2023mamba} with $N=4$ stacked Mamba blocks. Each block applies a Mamba mixer with residual connections and layer normalization in a pre-norm scheme. The input to the decoder is the concatenation of the visual and text sequences:
\begin{equation}
\mathbf{X} = [\mathbf{V}_{\text{enc}} + \mathbf{E}_{\text{seg}}[0]; \mathbf{T} + \mathbf{E}_{\text{seg}}[1]] \in \mathds{R}^{(L_v + L_t) \times 256}
\end{equation}
where $\mathbf{T}$ is the sequence of token embeddings. Each Mamba block uses state dimension 256 and expansion factor 6 (projecting the 256-dimensional input to an internal dimension of 1536 before the selective SSM operation). Causality is naturally enforced by the recurrent state of SSMs: each position can only access the past through the state, making explicit causal attention masks unnecessary. The output is projected to vocabulary logits by a linear head:
\begin{equation}
P(y_t | y_{<t}, \mathbf{I}) = \text{softmax}(\mathbf{W}_{\text{LM}} \cdot \mathbf{h}_t)
\end{equation}
where $\mathbf{h}_t$ is the $t$-th output of the decoder corresponding to a text position. Training uses teacher forcing with cross-entropy loss.

\subsubsection{Efficient Inference.}
\label{subsubsec:no-cross-attn}
A structural difference with Transformer-based recognizers such as DAN \cite{coquenet2023dan} or TrOCR \cite{li2021trocr} deserves emphasis. In those architectures, every decoded character can re-attend to the visual feature map via cross-attention and to previous characters via self-attention. In Mamba-OCR, the decoder does \emph{not} look back at the encoder output after the initial visual conditioning: each step consumes only the previous token embedding and updates the recurrent SSM state. All the information needed to transcribe the paragraph must therefore be compressed, once, into the connector output $\mathbf{V}_{\text{enc}} \in \mathds{R}^{L_v \times 256}$ and propagated through the SSM state, with $L_v$ scaling with the input image. At inference time, this enables a constant-memory decoding loop: the visual sequence and initial tokens initialize the SSM states in a single forward pass, after which each subsequent token is decoded one at a time with only the new token as input and incrementally updated states. There is no growing key-value cache: memory per decoding step is constant regardless of sequence length, a critical property for long-paragraph recognition. The trade-off is that the representational burden shifts from a flexible per-step query to a single, frozen visual summary.

\section{Methodology}
\label{sec:methodology}

\subsection{Synthetic Paragraph Generation}
\label{subsec:synth-gen}

To isolate the effect of sequence length from document-quality factors, we generate synthetic paragraphs from English Wikipedia text. Each paragraph is rendered on a white background, without degradation or layout variation. This controlled setting enables precise characterization of length-dependent behavior.

We use multiple target configurations with approximately 100 characters per line:
\begin{itemize}
    \item \textbf{Short:} 1 line ($\sim$100 chars), 3 lines ($\sim$300 chars)
    \item \textbf{Medium:} 5 lines ($\sim$500 chars), 7 lines ($\sim$700 chars)
    \item \textbf{Long:} 10 lines ($\sim$1000 chars)
\end{itemize}

This maximum matches the practical limit observed in \cite{agbeti2026benchmark} and the 1000-character cap used in the paragraph-level BnL subset (1--10 lines).

\subsubsection{Critical role of training data volume.}
An important observation emerged during preliminary experiments. With a \emph{fixed} set of 8,000 training paragraphs per length, Mamba-OCR failed to converge even on 2-line paragraphs ($\sim$200 chars): training loss decreased normally, but the model memorized training examples and validation loss plateaued. This data-hunger appears to be specific to the autoregressive Mamba decoder: the SSM-based AR head fuses the visual context and the previously decoded tokens into a single compressed recurrent state, and learning that fusion to generalize requires substantial lexical and syntactic diversity at training time.

We resolved this by generating $\sim$1M synthetic paragraphs per configuration, with text segments sampled dynamically from Wikipedia rather than from a fixed set. This order-of-magnitude increase in data diversity enabled stable convergence; we revisit this point in Section \ref{sec:realworld} when analyzing IAM results.

\subsection{Training Protocol}
\label{subsec:training-proto}

All experiments use identical training hyperparameters to ensure fair comparison:
\begin{itemize}
    \item Optimizer: AdamW, learning rate $10^{-4}$, weight decay $10^{-4}$
    \item Batch size: 4 (limited by paragraph image memory)
    \item Early stopping based on validation CER (patience 15 epochs)
    \item Data augmentation (probability 0.5): DPI rescaling, perspective distortion, elastic deformation, Gaussian blur/noise
\end{itemize}

Scaling results (Section \ref{sec:scaling}) and real-world results (Section \ref{sec:realworld}) are reported using the same protocol.

\subsection{Baseline and Metrics}
\label{subsec:baseline-metrics}

We compare Mamba-OCR against DAN \cite{coquenet2023dan}, the autoregressive Transformer baseline sharing the same CNN encoder, following \cite{agbeti2026benchmark}. DAN uses an 8-layer Transformer decoder with 4 attention heads and $\sim$7.1M parameters (vs 14.1M for Mamba-OCR at the recommended configuration).

\textbf{Identical training protocol for DAN and Mamba-OCR.} For a fair head-to-head comparison, DAN is retrained from scratch on the same synthetic datasets and with the same optimization and augmentation settings as Mamba-OCR (Sections \ref{subsec:synth-gen} and \ref{subsec:training-proto}). DAN uses the standard PyTorch \texttt{nn.MultiheadAttention}. Aside from the backbone (Transformer decoder vs Mamba decoder), no architectural or optimisation differences were introduced.

We report Character Error Rate (CER), inference latency (ms/image) on NVIDIA A100 with batch size 4, and parameter counts.

\section{Exploration of Hyperparameters}
\label{sec:ablation}

We now investigate how the four key architectural hyperparameters of Mamba-OCR, namely decoder depth, state dimension, expansion factor, and multimodal connector depth, affect paragraph-level recognition. Rather than running an exhaustive grid, we explore these hyperparameters in the order suggested by practical constraints (compute budget, inference latency, and the recurrent state's memory capacity). This section reports that exploration as it actually unfolded; the resulting recommendation, summarised in Section \ref{subsec:ablation-summary}, is consistent with and supports the defaults adopted by \cite{agbeti2026benchmark}.

\subsection{Setup and Search Strategy}
\label{subsec:ablation-setup}

\textbf{Starting point.} Our exploration starts from a deliberately small baseline configuration, $L=4$ decoder layers, state dimension $N=16$, expansion factor $E=2$, and $\text{MC}=1$ connector layer. Each subsequent step tries to improve over this baseline along a single axis at a time, holding the others fixed at their then-current best value.

\textbf{Search order.} The exploration proceeds in four stages, in the order in which they were run: (i) decoder depth $L$, the most natural way to add capacity but the one that most directly inflates inference latency; (ii) state dimension $N$, identified as the next lever once depth proved costly, since it adds memory \emph{without} lengthening the sequential decoding loop; (iii) expansion factor $E$, a complementary capacity lever that widens each Mamba block's internal projection; and (iv) connector depth $\text{MC}$, finally checked against the default of \cite{agbeti2026benchmark}.

\textbf{Length-aware early stopping.} Each configuration is trained on synthetic paragraphs of increasing length, but we stop investing compute in a configuration as soon as it is clearly dominated by a previously evaluated one at the shorter lengths. Cells corresponding to lengths at which a configuration was not trained are marked ``n/a''.

\textbf{Reporting.} CER is reported on the synthetic test set at $\sim$100~ch (1 line), $\sim$300~ch (3 lines), $\sim$500~ch (5 lines), $\sim$700~ch (7 lines), and $\sim$1000~ch (10 lines). Inference latency (ms) per image is given next to each CER value. All other training settings follow Section \ref{sec:methodology}.

\subsection{Decoder Depth}
\label{subsec:ablation-depth}

The first axis we explored was decoder depth $L$, with the state dimension and expansion factor held at the small baseline values ($N=16$, $E=2$, $\text{MC}=1$). We trained models with $L \in \{2, 4, 8, 16\}$ at one line as a fast capacity check, and extended only the most promising depths to longer sequences (Table \ref{tab:ablation-depth}).

\begin{table}[ht]
\small
\centering
\caption{Decoder depth at $N=16$, $E=2$, $\text{MC}=1$. CER (\%) and inference latency (Lat., ms) on synthetic paragraphs of increasing length. ``n/a'' = not measured (see Section \ref{subsec:ablation-setup}).}
\label{tab:ablation-depth}
\resizebox{\textwidth}{!}{%
\begin{tabular}{@{}lccccccccc@{}}
\toprule
& \multicolumn{2}{c}{\textbf{1 line}} & \multicolumn{2}{c}{\textbf{3 lines}} & \multicolumn{2}{c}{\textbf{5 lines}} & \multicolumn{2}{c}{\textbf{7 lines}} & \\
\cmidrule(lr){2-3}\cmidrule(lr){4-5}\cmidrule(lr){6-7}\cmidrule(lr){8-9}
\textbf{Depth} & \textbf{CER (\%)} & \textbf{Lat. (ms)} & \textbf{CER (\%)} & \textbf{Lat. (ms)} & \textbf{CER (\%)} & \textbf{Lat. (ms)} & \textbf{CER (\%)} & \textbf{Lat. (ms)} & \textbf{Params} \\
\midrule
$L=2$  & 6.14 & 53.1  & n/a  & n/a   & n/a  & n/a   & n/a  & n/a   & 3.2M \\
$L=4$  & 0.98 & 82.9  & 1.18 & 198.5 & 3.84 & 318.4 & 8.29 & 426.3 & 4.1M \\
$L=8$  & \textbf{0.14} & 125.6 & \textbf{0.44} & 334.3 & \textbf{1.25} & 542.2 & \textbf{6.19} & 761.2 & 5.8M \\
$L=16$ & 0.62 & 219.8 & 0.71 & 613.1 & 5.43 & 997.6 & n/a  & n/a   & 9.3M \\
\bottomrule
\end{tabular}%
}
\end{table}

Three observations emerge. First, $L=2$ collapses already at one line (6.14\% CER) and was not extended further. Second, increasing depth from $L=4$ to $L=8$ does improve CER at every measured length, but the per-image latency rises by roughly 50--80\% (e.g., from 318 ms to 542 ms at five lines). Third, $L=16$ is \emph{not} a monotone improvement: at five lines it reaches 5.43\% CER, worse than $L=8$ (1.25\%), while costing nearly 1 second of inference per image. This pattern points to \emph{recurrent-state capacity}, not decoder depth, as the binding constraint, and motivates the next stage of the search. We retain $L=4$ as a good accuracy/latency operating point for the rest of the exploration; we will return to $L=8$ when checking whether the conclusions about $N$ and $E$ hold at greater depth.

\subsection{State Dimension}
\label{subsec:ablation-state}

Having ruled out decoder depth alone as the primary capacity lever, we turned to the state dimension $N$, which controls the size of the recurrent state through which all past information must propagate. Larger $N$ adds memory \emph{without} lengthening the sequential decoding loop, so it can improve long-sequence accuracy without the latency penalty of deeper decoders. We fixed $E=2$, $\text{MC}=1$ and varied $N \in \{16, 32, 64, 128, 256\}$ at $L=4$, and additionally cross-checked $N \in \{16, 128, 256\}$ at $L=8$ to ensure that the conclusions are not specific to one depth (Table \ref{tab:ablation-state}). The upper bound $N=256$ is the largest state dimension compatible with the inference cache and parallel-scan throughput of the \texttt{mamba-ssm} kernels in our implementation; values beyond this point either exhausted GPU memory or broke parallel-scan throughput on a single GPU.

\begin{table}[ht]
\small
\centering
\caption{State dimension at $E=2$, $\text{MC}=1$. CER (\%) and inference latency (Lat., ms) on synthetic paragraphs of increasing length.}
\label{tab:ablation-state}
\resizebox{\textwidth}{!}{%
\begin{tabular}{@{}llcccccccccccc@{}}
\toprule
& & \multicolumn{2}{c}{\textbf{1 line}} & \multicolumn{2}{c}{\textbf{3 lines}} & \multicolumn{2}{c}{\textbf{5 lines}} & \multicolumn{2}{c}{\textbf{7 lines}} & \multicolumn{2}{c}{\textbf{10 lines}} & \\
\cmidrule(lr){3-4}\cmidrule(lr){5-6}\cmidrule(lr){7-8}\cmidrule(lr){9-10}\cmidrule(lr){11-12}
\textbf{Depth} & \textbf{State} & \textbf{CER (\%)} & \textbf{Lat. (ms)} & \textbf{CER (\%)} & \textbf{Lat. (ms)} & \textbf{CER (\%)} & \textbf{Lat. (ms)} & \textbf{CER (\%)} & \textbf{Lat. (ms)} & \textbf{CER (\%)} & \textbf{Lat. (ms)} & \textbf{Params} \\
\midrule
$L=4$ & $N=16$  & 0.98 & 82.9  & 1.18 & 198.5 & 3.84 & 318.4 & 8.29 & 426.3 & n/a   & n/a    & 4.1M \\
$L=4$ & $N=32$  & 0.69 & 89.2  & 1.47 & 204.6 & 4.17 & 324.1 & n/a  & n/a   & n/a   & n/a    & 4.2M \\
$L=4$ & $N=64$  & 0.66 & 93.4  & 0.57 & 207.6 & 2.61 & 326.7 & n/a  & n/a   & n/a   & n/a    & 4.4M \\
$L=4$ & $N=128$ & 0.89 & 92.4  & 0.96 & 208.1 & 1.41 & 332.1 & 11.80 & 451.3 & n/a  & n/a    & 4.9M \\
$L=4$ & $N=256$ & 0.57 & 100.1 & 0.65 & 213.2 & 1.87 & 332.1 & 3.61 & 455.8 & 5.28  & 644.4  & 5.9M \\
\midrule
$L=8$ & $N=16$  & 0.14 & 125.6 & 0.44 & 334.3 & 1.25 & 542.2 & 6.19 & 761.2 & n/a   & n/a    & 5.8M \\
$L=8$ & $N=128$ & 0.73 & 145.2 & 0.31 & 350.4 & 0.53 & 565.5 & 2.11 & 782.2 & 10.17 & 1104.6 & 7.3M \\
$L=8$ & $N=256$ & \textbf{0.24} & 159.7 & \textbf{0.49} & 366.5 & \textbf{0.82} & 586.3 & \textbf{1.41} & 810.6 & \textbf{2.07} & 1153.1 & 9.1M \\
\bottomrule
\end{tabular}%
}
\end{table}

Two patterns emerge. First, at fixed depth $L=4$, increasing $N$ from 16 to 256 monotonically improves CER once sequences become long: the gain is small at one line but substantial at five lines (3.84\%~$\rightarrow$~1.87\%) and decisive at seven lines (8.29\%~$\rightarrow$~3.61\%). Even more strikingly, $N=128$ at $L=4$, $7$ lines collapses to 11.80\%, while $N=256$ at the same depth stays at 3.61\%, suggesting that a slightly undersized state can fail catastrophically rather than degrade gracefully. Second, the same conclusion holds at $L=8$: $N=256$ dominates $N=128$ and $N=16$ at every length, and is the only configuration that reaches a sub-3\% CER at 10 lines (2.07\%). In neither sweep did we observe the kind of plateau or degradation that we will see for excessively large expansion factors in Section \ref{subsec:ablation-expansion}; every increase in $N$ either helped or, at worst, did not hurt at the lengths we could measure.

Taken together, these observations identify $N=256$ as a \emph{plateau forced by implementation constraints} rather than an accuracy peak: lifting the cache and parallel-scan limits would likely allow larger state dimensions to further improve long-paragraph accuracy. We adopt \textbf{$N=256$} as the new default for the remainder of the search, with this caveat in mind.

\subsection{Expansion Factor}
\label{subsec:ablation-expansion}

With $N$ fixed at $256$, we then explored the expansion factor $E$, which sets the internal width of each Mamba block (from $D=256$ to $D \times E$). Expansion is a complementary capacity lever to the state dimension: it widens the per-block computation rather than the recurrent memory. We tested $E \in \{2, 6, 10, 12, 20\}$ at $L=4$, $N=256$, $\text{MC}=1$, with $E=2$ and $E=6$ trained up to 10 lines, and the larger values evaluated more cheaply at up to 5 lines as a screening step. We also recall the corresponding $L=8$, $N=256$, $E=2$ row from Table \ref{tab:ablation-state} for context (Table \ref{tab:ablation-expansion}).

\begin{table}[ht]
\small
\centering
\caption{Expansion factor (Exp) at $L=4$ (and one $L=8$ reference row), $N=256$, $\text{MC}=1$. CER (\%) and inference latency (Lat., ms) on synthetic paragraphs of increasing length.}
\label{tab:ablation-expansion}
\resizebox{\textwidth}{!}{%
\begin{tabular}{@{}llcccccccccccc@{}}
\toprule
& & \multicolumn{2}{c}{\textbf{1 line}} & \multicolumn{2}{c}{\textbf{3 lines}} & \multicolumn{2}{c}{\textbf{5 lines}} & \multicolumn{2}{c}{\textbf{7 lines}} & \multicolumn{2}{c}{\textbf{10 lines}} & \\
\cmidrule(lr){3-4}\cmidrule(lr){5-6}\cmidrule(lr){7-8}\cmidrule(lr){9-10}\cmidrule(lr){11-12}
\textbf{Depth} & \textbf{Exp} & \textbf{CER (\%)} & \textbf{Lat. (ms)} & \textbf{CER (\%)} & \textbf{Lat. (ms)} & \textbf{CER (\%)} & \textbf{Lat. (ms)} & \textbf{CER (\%)} & \textbf{Lat. (ms)} & \textbf{CER (\%)} & \textbf{Lat. (ms)} & \textbf{Params} \\
\midrule
$L=4$ & $E=2$  & 0.57 & 100.1 & 0.65 & 213.2 & 1.87 & 332.1 & 3.61 & 455.8 & 5.28 & 644.4 & 5.9M \\
$L=4$ & $E=6$  & 0.48 & 99.7  & 0.74 & 245.4 & \textbf{0.59} & 396.4 & \textbf{0.22} & 556.9 & \textbf{0.98} & 774.9 & 14.1M \\
$L=4$ & $E=10$ & \textbf{0.15} & 116.7 & \textbf{0.18} & 277.2 & 0.61 & 444.8 & n/a  & n/a   & n/a  & n/a    & 22.1M \\
$L=4$ & $E=12$ & \textbf{0.15} & 123.1 & 0.17 & 301.7 & 1.33 & 469.9 & n/a  & n/a   & n/a  & n/a    & 26.1M \\
$L=4$ & $E=20$ & 0.23 & 155.6 & 0.31 & 368.9 & 0.72 & 582.1 & n/a  & n/a   & n/a  & n/a    & 42.2M \\
\midrule
$L=8$ & $E=2$  & 0.24 & 159.7 & 0.49 & 366.5 & 0.82 & 586.3 & 1.41 & 810.6 & 2.07 & 1153.1 & 9.1M \\
\bottomrule
\end{tabular}%
}
\end{table}

Two findings stand out. First, moving from $E=2$ to $E=6$ at $L=4$ produces a substantial improvement that grows with sequence length: from a near-tie at one line (0.48\% vs 0.57\%) to a $5\times$ reduction at 10 lines (0.98\% vs 5.28\% CER). This is the best long-sequence configuration in our exploration, surpassing even $L=8$, $E=2$ (2.07\% at 10 lines, but at $1.5\times$ the latency at one line and twice the parameters of the encoder--decoder stack).
Second, going beyond $E=6$ does not pay off. Although $E=10$ and $E=12$ reach excellent CER at very short sequences (0.15--0.18\% at one and three lines), they were not extended past five lines and already show signs of irregular behaviour at that length (e.g., $E=12$ jumps to 1.33\% CER at 5 lines, well above $E=6$'s 0.59\%). $E=20$, despite a parameter count of 42.2M, fails to improve over $E=6$ even at the lengths we did measure, while costing roughly $50\%$ more inference time. We attribute this reversal to an optimisation difficulty: excessively wide internal projections may cause mild training instabilities or overfitting when training data is finite, an explanation consistent with the data-scarcity discussion of Section \ref{sec:realworld}. We retain \textbf{$E=6$} as the best expansion factor at $L=4$.

\subsection{Multimodal Connector Depth}
\label{subsec:ablation-connector}
Finally, we verified that the default single-layer bidirectional Mamba connector used in \cite{agbeti2026benchmark} remains appropriate in our setting. A connector layer establishes global visual context before decoding by running a Mamba block forward and backward over the flattened CNN feature map (Section \ref{subsec:mamba-ocr-arch}). Our partial connector data points, collected as a side observation rather than as a controlled one-axis sweep, were consistent with the conclusions of \cite{agbeti2026benchmark}: removing the connector entirely ($\text{MC}=0$) noticeably hurt accuracy, while configurations with $\text{MC}>1$ added latency without a correspondingly clear accuracy improvement at the lengths we could measure. We therefore retain $\text{MC}=1$, in agreement with \cite{agbeti2026benchmark}, and report this finding as a \emph{consistency check} rather than as a controlled ablation, leaving a fuller MC sweep at $L=4$, $N=256$, $E=6$ to future work.

\subsection{Recommended Configuration}
\label{subsec:ablation-summary}

Aggregating the four stages of the exploration, the configuration that gave the best results in our search is:
\begin{itemize}
    \item \textbf{Decoder depth:} $L=4$ Mamba layers. Depth above this point (i.e., $L=8$) yields clearly better CER at the small state dimension $N=16$, but the picture inverts once $N$ and $E$ are properly chosen: at $L=4$, $N=256$, $E=6$ the model reaches 0.98\% CER at 10 lines at roughly half the inference cost of the best $L=8$ setting (2.07\% CER, 1153 ms).
    \item \textbf{State dimension:} $N=256$, the maximum allowed by cache size and training parallelism in our implementation. Together with expansion factor, this was one of the two most rewarding levers; larger values would likely help further if hardware constraints could be relaxed.
    \item \textbf{Expansion factor:} $E=6$, the best of the values $\{2, 6, 10, 12, 20\}$ tested at $L=4$, $N=256$. Both smaller and larger values were dominated for long sequences, with $E \geq 10$ exhibiting both larger parameter counts and signs of optimisation instability.
    \item \textbf{Multimodal connector depth:} $\text{MC}=1$, consistent with the default \cite{agbeti2026benchmark}.
\end{itemize}

This configuration coincides with the default adopted in \cite{agbeti2026benchmark}. Given the partial coverage of our search, we describe this agreement as empirical \emph{support} for that configuration in our setting, rather than as a formal ablation-based justification: among the configurations we were able to train, none outperformed $(L, N, E, \text{MC}) = (4, 256, 6, 1)$ on paragraph-level synthetic data up to 10 lines. The state dimension is the only hyperparameter for which our chosen value is bounded by implementation constraints rather than by an accuracy plateau, an important nuance when interpreting the scaling results that follow. All subsequent experiments use this configuration.

\section{Scaling Study}
\label{sec:scaling}

With the recommended configuration established, we now characterize how Mamba-OCR scales with sequence length and compare it against DAN on the same controlled synthetic data.

\subsection{Scaling Curves}
\label{subsec:scaling-curves}

\begin{figure}[ht]
\centering
\begin{tikzpicture}
\begin{axis}[
    width=0.85\textwidth,
    height=6cm,
    xlabel={Sequence Length (characters)},
    ylabel={Inference Latency (ms)},
    xmin=0, xmax=1100,
    ymin=0, ymax=4000,
    legend pos=north west,
    grid=major,
    legend style={font=\small}
]
\addplot[color=blue, mark=square*, thick] coordinates {
    (100, 137.1) (300, 373.6) (500, 918.9) (700, 1656.6) (1000, 3511.2)
};
\addplot[color=red, mark=triangle*, thick] coordinates {
    (100, 99.7) (300, 245.4) (500, 396.4) (700, 556.9) (1000, 774.9)
};
\legend{DAN, Mamba-OCR}
\end{axis}
\end{tikzpicture}
\caption{Inference latency scaling with sequence length on clean synthetic paragraphs. Mamba-OCR consistently exhibits lower latency; the speedup over DAN grows with sequence length (see Table~\ref{tab:scaling} for exact ratios).}
\label{fig:scaling}
\end{figure}
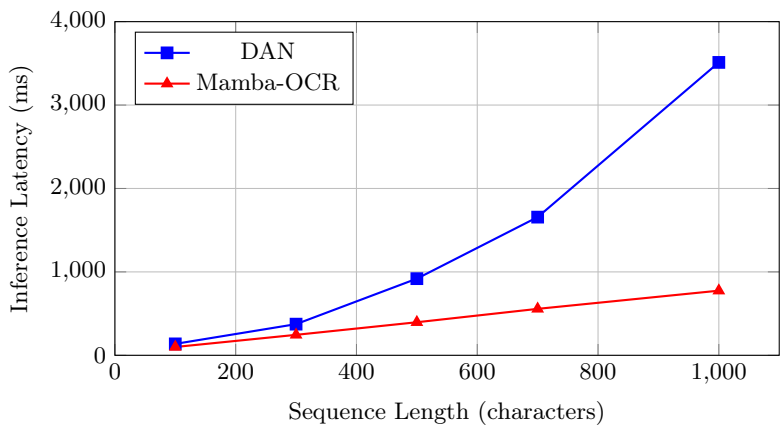

Table \ref{tab:scaling} and Figure \ref{fig:scaling} present CER and per-image inference latency as a function of sequence length, with DAN at the configuration of \cite{coquenet2023dan} and Mamba-OCR at the recommended setting ($L=4$, $N=256$, $E=6$, $\text{MC}=1$).

\begin{table}[ht]
\small
\centering
\caption{Scaling results on clean synthetic paragraphs. CER (\%) and inference latency (Lat., ms) for DAN and the recommended Mamba-OCR configuration ($L=4$, $N=256$, $E=6$, $\text{MC}=1$).}
\label{tab:scaling}
\resizebox{\textwidth}{!}{%
\begin{tabular}{@{}cccccccc@{}}
\toprule
& & \multicolumn{2}{c}{\textbf{DAN}} & \multicolumn{2}{c}{\textbf{Mamba-OCR}} & & \\
\cmidrule(lr){3-4}\cmidrule(lr){5-6}
\textbf{Lines} & \textbf{Chars} & \textbf{CER (\%)} & \textbf{Lat. (ms)} & \textbf{CER (\%)} & \textbf{Lat. (ms)} & \textbf{Rel. Gap} & \textbf{Speedup} \\
\midrule
1  & $\sim$100  & 0.93 & 137.1  & \textbf{0.48} & \textbf{99.7}  & $-$48\% & 1.38$\times$ \\
3  & $\sim$300  & \textbf{0.13} & 373.6  & 0.74 & \textbf{245.4} & +469\% & 1.52$\times$ \\
5  & $\sim$500  & 0.76 & 918.9  & \textbf{0.59} & \textbf{396.4} & $-$22\% & 2.32$\times$ \\
7  & $\sim$700  & 0.31 & 1656.6 & \textbf{0.22} & \textbf{556.9} & $-$29\% & 2.97$\times$ \\
10 & $\sim$1000 & \textbf{0.21} & 3511.2 & 0.98 & \textbf{774.9} & +367\% & 4.53$\times$ \\
\bottomrule
\end{tabular}%
}
\end{table}

Three observations emerge.
\begin{itemize}
    \item \textbf{Both models perform well on clean synthetic data, and neither dominates in accuracy.} CER stays under 1\% at every measured length for both architectures, well below the 2\%--6\% range seen on real BnL paragraphs \cite{agbeti2026benchmark}. Mamba-OCR is better at 1, 5, and 7 lines (0.48\%, 0.59\%, 0.22\% vs DAN's 0.93\%, 0.76\%, 0.31\%), while DAN is better at 3 and 10 lines (0.13\%, 0.21\% vs Mamba-OCR's 0.74\%, 0.98\%). Both architectures stay within roughly $\pm$0.8\% absolute CER of each other, with no systematic accuracy winner across lengths.
    \item \textbf{Mamba-OCR's efficiency advantage grows steeply with length.} The per-image speedup over DAN moves from $1.38\times$ at 1 line to $4.53\times$ at 10 lines (i.e., 775 ms vs 3511 ms), reflecting the $O(n)$ vs $O(n^2)$ complexity gap. This is the most consistent and most actionable finding of the synthetic scaling study: in the absence of a clear accuracy winner, Mamba-OCR processes long paragraphs several times faster than DAN at essentially the same error rate, in line with the prior efficiency results of \cite{agbeti2026benchmark}.
    \item \textbf{The synthetic regime is the favourable case.} The fact that both architectures comfortably stay below 1\% CER on every length here should be contrasted with the more nuanced picture on real documents (Section \ref{sec:realworld}), where document-level noise, layout variability, and especially limited training data on IAM substantially change the conclusions.
\end{itemize}

\section{Real-World Validation and Limitations}
\label{sec:realworld}

The controlled synthetic study characterizes Mamba-OCR's intrinsic behavior on clean, well-resourced data. We now examine how these findings transfer to real documents across four settings: printed and handwritten text, at both line and paragraph granularities. Our goal is to assess when Mamba-OCR holds its promise and when it reaches its limits.

\subsection{Printed Lines (BnL Antiqua and Fraktur)}
\label{subsec:real-bnl-lines}

We first recall the line-level results reported in \cite{agbeti2026benchmark} for printed historical newspapers. On the BnL line-level test set (8,328 lines across Antiqua and Fraktur), Mamba-OCR matches or comes very close to the strongest Transformer baselines. Table \ref{tab:bnl-lines-recall} summarizes the neural models.

\begin{table}[ht!]
\small
\centering
\begin{minipage}{0.48\textwidth}
\centering
\caption{BnL line-level results (printed), recalled from \cite{agbeti2026benchmark}. CER and WER (\%) on Antiqua and Fraktur test sets.}
\label{tab:bnl-lines-recall}
\resizebox{\textwidth}{!}{
\begin{tabular}{@{}lcccc@{}}
\toprule
& \multicolumn{2}{c}{\textbf{Antiqua}} & \multicolumn{2}{c}{\textbf{Fraktur}} \\
\cmidrule(lr){2-3} \cmidrule(lr){4-5}
\textbf{Model} & \textbf{CER} & \textbf{WER} & \textbf{CER} & \textbf{WER} \\
\midrule
VAN \cite{coquenet2023van}      & 1.85          & \textbf{3.39} & \textbf{3.01} & 5.21          \\
DAN \cite{coquenet2023dan}      & \textbf{1.83} & 3.61          & 3.03          & \textbf{4.88} \\
DANIEL \cite{constum2025daniel} & 2.37          & 4.99          & 6.18          & 14.11         \\
\textbf{Mamba-OCR}              & \textbf{1.83} & 3.84          & 3.06          & 5.73          \\
\bottomrule
\end{tabular}
}
\end{minipage}%
\hfill
\begin{minipage}{0.48\textwidth}
\centering
\caption{BnL paragraph-level results (printed, 1--10 lines), recalled from \cite{agbeti2026benchmark}. CER and WER (\%) on the BnL paragraph test set.}
\label{tab:bnl-para-recall}
\resizebox{\textwidth}{!}{
\begin{tabular}{@{}lcc@{}}
\toprule
\textbf{Model} & \textbf{CER} & \textbf{WER} \\
\midrule
VAN \cite{coquenet2023van}      & 6.42          & \textbf{7.64} \\
DAN \cite{coquenet2023dan}      & \textbf{5.24} & 9.14          \\
DANIEL \cite{constum2025daniel} & 6.18          & 9.53          \\
\textbf{Mamba-OCR}              & 6.07          & 8.81          \\
\bottomrule
\end{tabular}
}
\end{minipage}
\end{table}

On printed lines, Mamba-OCR is essentially on par with the best Transformer baselines: matching DAN exactly on Antiqua CER (1.83\%) and within 0.03\% on Fraktur (3.06\% vs 3.03\%). DANIEL's collapse on Fraktur (6.18\%) illustrates how BPE tokenization can fail for scripts poorly covered by the pretrained vocabulary, whereas character-level Mamba-OCR remains robust. These results confirm that SSMs are a fully viable choice for line-level recognition on printed historical text.

\subsection{Handwritten Lines (IAM)}
\label{subsec:real-iam-lines}

We then extend the evaluation to handwritten lines using the IAM Handwriting Database \cite{marti2002iam}, the standard benchmark for offline English handwriting recognition. Following the standard Aachen split (6,161 train / 940 val / 1,861 test lines), we train Mamba-OCR from scratch on IAM and compare against an extensive set of published baselines spanning CTC-based, attention-based, Transformer-based, and the recent RetNet-based DRetHTR \cite{kim2026drethtr}. Table \ref{tab:iam-lines} reports CER on the IAM test set.

\begin{table}[ht]
\small
\centering
\caption{Handwritten line-level recognition on IAM test set (Aachen split). Results are taken from the original papers. Models are grouped by architectural family. CER (\%) and WER (\%) reported where available.}
\label{tab:iam-lines}
\begin{tabular}{@{}llcc@{}}
\toprule
\textbf{Model} & \textbf{Architecture} & \textbf{CER (\%)} & \textbf{WER (\%)} \\
\midrule
\multicolumn{4}{l}{\textit{CTC-based}} \\
Voigtlaender et al.\ \cite{voigtlaender2016mdlstm} & MDLSTM + CTC & 3.5 & 9.3 \\
Puigcerver \cite{puigcerver2017} & CNN + BiLSTM + CTC & 5.8 & 18.4 \\
Bluche \& Messina \cite{bluche2017gru} & Gated CRNN + CTC & 3.2 & 10.5 \\
VAN \cite{coquenet2023van} & FCN + Hybrid Attn.\ + LSTM + CTC & 4.97 & 16.31 \\
\midrule
\multicolumn{4}{l}{\textit{Attention / Autoregressive}} \\
DAN \cite{coquenet2023dan} & FCN + Transformer AR & 4.23 & --- \\
\midrule
\multicolumn{4}{l}{\textit{Foundation / Pretrained models}} \\
TrOCR-small \cite{li2021trocr} & ViT + Transformer (fine-tuned) & 4.22 & --- \\
TrOCR-base \cite{li2021trocr} & ViT + Transformer (fine-tuned) & 3.42 & --- \\
TrOCR-large \cite{li2021trocr} & ViT + Transformer (fine-tuned) & 2.89 & --- \\
\midrule
\multicolumn{4}{l}{\textit{Retentive Network}} \\
DRetHTR\textsubscript{SMALL} \cite{kim2026drethtr} & RetNet decoder-only & 2.97 & 9.13 \\
DRetHTR\textsubscript{BASE} \cite{kim2026drethtr} & RetNet decoder-only & \textbf{2.26} & \textbf{6.55} \\
\midrule
\multicolumn{4}{l}{\textit{State-Space (ours)}} \\
\textbf{Mamba-OCR} & CNN + BiMamba + AR Mamba & 8.21 & 17.39 \\
\bottomrule
\end{tabular}
\end{table}

On IAM lines, Mamba-OCR reaches \textbf{8.21\% CER}, clearly below all baselines we compare to. It is roughly $2\times$ worse than DAN (4.23\%) and TrOCR-small (4.22\%), and more than $3\times$ worse than the best RetNet-based DRetHTR\textsubscript{BASE} (2.26\%). The contrast with the Retentive Network family is particularly informative: DRetHTR is also a linear-time autoregressive backbone, but it is a decoder-only model trained at a different scale and with a different recipe, and it currently sets the state of the art on this benchmark.

This line-level IAM result already tells an important story: at the line level, where data is reasonably abundant (6,161 training lines), Mamba-OCR generalizes from printed to handwritten text but is no longer competitive with the strongest baselines. The contrast with the printed setting (BnL lines, where Mamba-OCR is on par with DAN) and the synthetic setting (Section \ref{sec:scaling}, where Mamba-OCR is sub-1\% across all lengths) suggests that the inductive bias gap, partly visible on IAM lines, becomes much more severe at paragraph level as we will see next.

\subsection{Printed Paragraphs (BnL)}
\label{subsec:real-bnl-para}
On BnL paragraphs (1--10 lines, up to $\sim$1000 chars), Mamba-OCR remains competitive Table \ref{tab:bnl-para-recall} \cite{agbeti2026benchmark}: 6.07\% CER versus DAN's 5.24\% (+15.8\% relative), with VAN at 6.42\%, DANIEL at 6.18\%, and Gemini at 6.06\%, while running 2.05$\times$ faster than DAN (195.6 ms vs 401.2 ms). Compared to the synthetic regime of Section \ref{sec:scaling}, document-level noise and layout variability erode Mamba-OCR's accuracy advantage, leaving DAN slightly ahead in absolute CER while Mamba-OCR retains the efficiency lead.

\subsection{Handwritten Paragraphs (IAM)}
\label{subsec:real-iam-para}

The most challenging setting combines handwriting with paragraph-level sequence lengths. We evaluate on IAM paragraphs \cite{marti2002iam} using the standard split (747 train / 116 val / 232 test paragraphs, average 450 characters).

\begin{table}[ht]
\small
\centering
\caption{IAM paragraph recognition (avg 450 characters).}
\label{tab:iam-para}
\begin{tabular}{@{}lccccc@{}}
\toprule
\textbf{Model} & \textbf{CER (\%)} & \textbf{WER (\%)} & \textbf{Latency} & \textbf{Speedup} \\
\midrule
DAN \cite{coquenet2023dan} & \textbf{3.51} & \textbf{10.23} & 385 ms & 1.0$\times$ \\
Mamba-OCR & 10.02 & 26.78 & 178 ms & 2.16$\times$ \\
\midrule
\textit{Relative gap} & +185\% & +162\% & -- & -- \\
\bottomrule
\end{tabular}
\end{table}

Table \ref{tab:iam-para} reveals the most substantial gap of our study: Mamba-OCR achieves 10.02\% CER versus DAN's 3.51\%, a \textbf{+185\% relative gap}. This is larger than the gap observed on BnL paragraphs (+16\%) and on IAM lines (+94\% vs DAN's 4.23\%). The combination of (i) handwriting difficulty and (ii) paragraph-level sequence length produces a \emph{compounding} effect rather than a purely additive one: each factor individually is tolerable, but together they push Mamba-OCR outside its operating regime.

\subsection{Is the IAM Gap Architectural or Data-Driven?}
\label{subsec:data-scarcity}

Such large gaps might suggest an architectural limitation of Mamba, but our synthetic experiments suggest that a significant portion is better explained by data scarcity than by architecture alone.

\textbf{Observation 1: Overfitting with limited data.}
With a fixed set of 8,000 training paragraphs per length, the autoregressive Mamba-OCR failed to converge even on 2-line paragraphs ($\sim$200 chars): training loss decreased normally, but the model memorized training examples and validation loss plateaued. DAN, in the same setting, showed less severe symptoms, indicating that SSMs' compressed recurrent state requires high lexical and structural diversity to generalize.

\textbf{Observation 2: Convergence with large-scale data.}
Moving to 1 million dynamically-sampled paragraphs per configuration resolved the issue entirely: validation loss tracked training loss closely, and Mamba-OCR reached the strong CERs of Section \ref{sec:scaling}.

\textbf{Observation 3: IAM falls in the data-scarce regime.}
IAM's 747 training paragraphs sit roughly an order of magnitude below the 8K threshold at which convergence failed in our synthetic setup. Mixing synthetic handwriting-like paragraphs during fine-tuning did not close the gap. The line-level result (8.21\% CER on 6,161 lines) is consistent: more data than for paragraphs, but still far from the synthetic scale at which Mamba-OCR converged. Notably, the RetNet-based DRetHTR \cite{kim2026drethtr} reaches 2.26\% on the same line split, showing that linear-time backbones \emph{are} compatible with strong IAM performance: the recipe and pretraining regime matter substantially.

\subsubsection{An architectural perspective on the data hunger.}
The data-hunger we observe also has a clean architectural reading (Section \ref{subsubsec:no-cross-attn}). Because the Mamba decoder never re-attends to the image, all visual evidence must be packed once into the encoder output and propagated through the SSM state. This compression buys Mamba its linear-time decoding and $2\times$--$4.5\times$ speed advantage, but it concentrates the learning burden on a single up-front representation that must support all subsequent decoding steps. Training such a representation to generalize requires far more data than a Transformer that can re-query the image on demand, consistent with our 8K-vs-1M observation on synthetic and with the persistent IAM gap below that scale. It also explains, symmetrically, why DRetHTR \cite{kim2026drethtr}, also a linear-time AR backbone, can nevertheless reach 2.26\% on IAM lines: its decoder-only design and pretraining regime address the same constraint through a different recipe.

\subsection{Practical Implications}
\label{subsec:practical}

Our findings refine the practical guidance from \cite{agbeti2026benchmark}:

\begin{itemize}
    \item \textbf{Mamba-OCR is well-suited for} printed document digitization at scale (lines and paragraphs), where large-scale synthetic pre-training is feasible. The 2$\times$--3$\times$ speedup over Transformer baselines translates directly to throughput gains in digitization workflows.
    \item \textbf{The autoregressive Mamba decoder is currently weaker on handwritten lines} (8.21\% CER on IAM) than the strongest Transformer baselines (TrOCR-large 2.89\%, DAN 4.23\%) and the recent RetNet-based DRetHTR (2.26\%), indicating that the inductive bias gap is already visible at the line level on this domain.
    \item \textbf{Mamba-OCR faces significant challenges on handwritten paragraphs} when training data is limited ($<$10K paragraphs), where Transformer baselines like DAN remain clearly preferable due to their stronger inductive bias on small datasets.
    \item \textbf{Training data volume is a critical hyperparameter} for SSM-based OCR. Future deployments on novel domains (new languages, new scripts, new handwriting styles) should budget for large-scale synthetic pre-training to unlock Mamba-OCR's full potential.
\end{itemize}

These observations point to clear future directions: (i) SSM-specific pre-training strategies on large synthetic handwriting corpora, (ii) hybrid SSM-attention architectures combining efficient processing with stronger inductive biases, and (iii) exploration of related linear-time backbones such as RetNet \cite{sun2023retnet,kim2026drethtr} and Gated DeltaNet \cite{yang2024gateddeltanet}, which share Mamba's efficiency goals but use different state-management mechanisms.

\section{Conclusion}
\label{sec:conclusion}

This paper extends \cite{agbeti2026benchmark} through a focused study of Mamba-OCR scaling from lines to paragraphs. A four-stage hyperparameter exploration over decoder depth ($L \in \{2,4,8,16\}$), state dimension ($N \in \{16, 32, 64, 128, 256\}$), expansion factor ($E \in \{2, 6, 10, 12, 20\}$), and multimodal connector depth supports the configuration of \cite{agbeti2026benchmark} ($L=4$, $N=256$, $E=6$, $\text{MC}=1$): no other configuration we trained outperformed it on paragraph-level synthetic data. The state dimension and the expansion factor emerged as the two most rewarding levers, and $N=256$ is the largest allowed by cache size and parallel-scan throughput in our implementation rather than an accuracy plateau, suggesting that further gains would be possible if these constraints could be relaxed. Controlled scaling on synthetic Wikipedia paragraphs shows that, with the same training data and protocol for both models, neither dominates in accuracy (both stay below 1\% CER up to 1000 characters), while Mamba-OCR remains 1.4--4.5$\times$ faster than DAN.

Real-world validation yields a more nuanced picture. On BnL printed paragraphs, Mamba-OCR is competitive (6.07\% vs 5.24\% for DAN) with a 2.05$\times$ speedup. On IAM handwritten lines, Mamba-OCR reaches 8.21\% CER, well behind both Transformer baselines (DAN 4.23\%, TrOCR-large 2.89\%) and DRetHTR (2.26\%). On IAM paragraphs, a substantial gap emerges (10.02\% vs 3.51\%, +185\%). Our analysis attributes a significant part of these gaps to data scarcity: in our controlled experiments, 8K fixed paragraphs were insufficient for the autoregressive Mamba decoder to converge on long sequences, while 1M dynamically-sampled paragraphs sufficed. IAM provides only 747 training paragraphs (and 6,161 lines), which falls within or close to the regime where convergence failed in our setup. SSMs thus offer compelling efficiency for printed digitization at scale but require either substantially more training data or architectural augmentation, such as hybrid SSM-attention designs or alternative linear-time backbones like DRetHTR \cite{kim2026drethtr} and Gated DeltaNet \cite{yang2024gateddeltanet}, for competitive handwriting recognition. Future work should explore SSM-specific pre-training on large synthetic handwriting corpora, hybrid Mamba-attention architectures, and scaling-aware training that exposes models to length and style variation.

\bibliographystyle{splncs04}

\end{document}